\documentclass[10pt,twocolumn,letterpaper]{article}

\usepackage{kotex}
\usepackage{afterpage}
\usepackage{float}
\usepackage{multirow}
\usepackage{algorithm}
\usepackage{algpseudocode}
\usepackage{amsmath}
\usepackage{arydshln}
\usepackage{colortbl}
\usepackage{longtable}
\usepackage{tabularx}
\usepackage{supertabular} 
\usepackage[accsupp]{axessibility}  

\usepackage[pagenumbers]{iccv} 

\definecolor{iccvblue}{rgb}{0.21,0.49,0.74}
\usepackage[pagebackref,breaklinks,colorlinks,allcolors=iccvblue]{hyperref}
\hypersetup{
  pdftitle={Know "No" Better: A Data-Driven Approach for Enhancing Negation Awareness in CLIP}
}

\def\paperID{8434} 
\def\confName{ICCV}
\def\confYear{2025}

\title{Know ``No'' Better:\\A Data-Driven Approach for Enhancing Negation Awareness in CLIP}

\author{
  Junsung Park$^1$ \hspace{0.2em}
  Jungbeom Lee$^{2}$ \hspace{0.2em}
  Jongyoon Song$^3$ \hspace{0.2em}
  Sangwon Yu$^1$ \hspace{0.2em}
  Dahuin Jung$^4$\textsuperscript{†} \hspace{0.2em}
  Sungroh Yoon$^{1,5}$\textsuperscript{†}
  \vspace{0.3em} \\
  $^1$Department of Electrical and Computer Engineering, Seoul National University \hspace{0.2em} $^2$Amazon \\
  $^3$Samsung Research \hspace{0.2em} $^4$School of Computer Science and Engineering, Soongsil University \\
  $^5$IPAI, AIIS, ASRI, INMC, and ISRC, Seoul National University \\
  \small \url{https://github.com/parkquasar/NegationCLIP}
  }

\begin{document}

\twocolumn[{%
\renewcommand\twocolumn[1][]{#1}%
\maketitle
\begin{center}
    \centering
    \captionsetup{type=figure}
    \vspace{-0.3em}
    \includegraphics[width=0.82\linewidth]{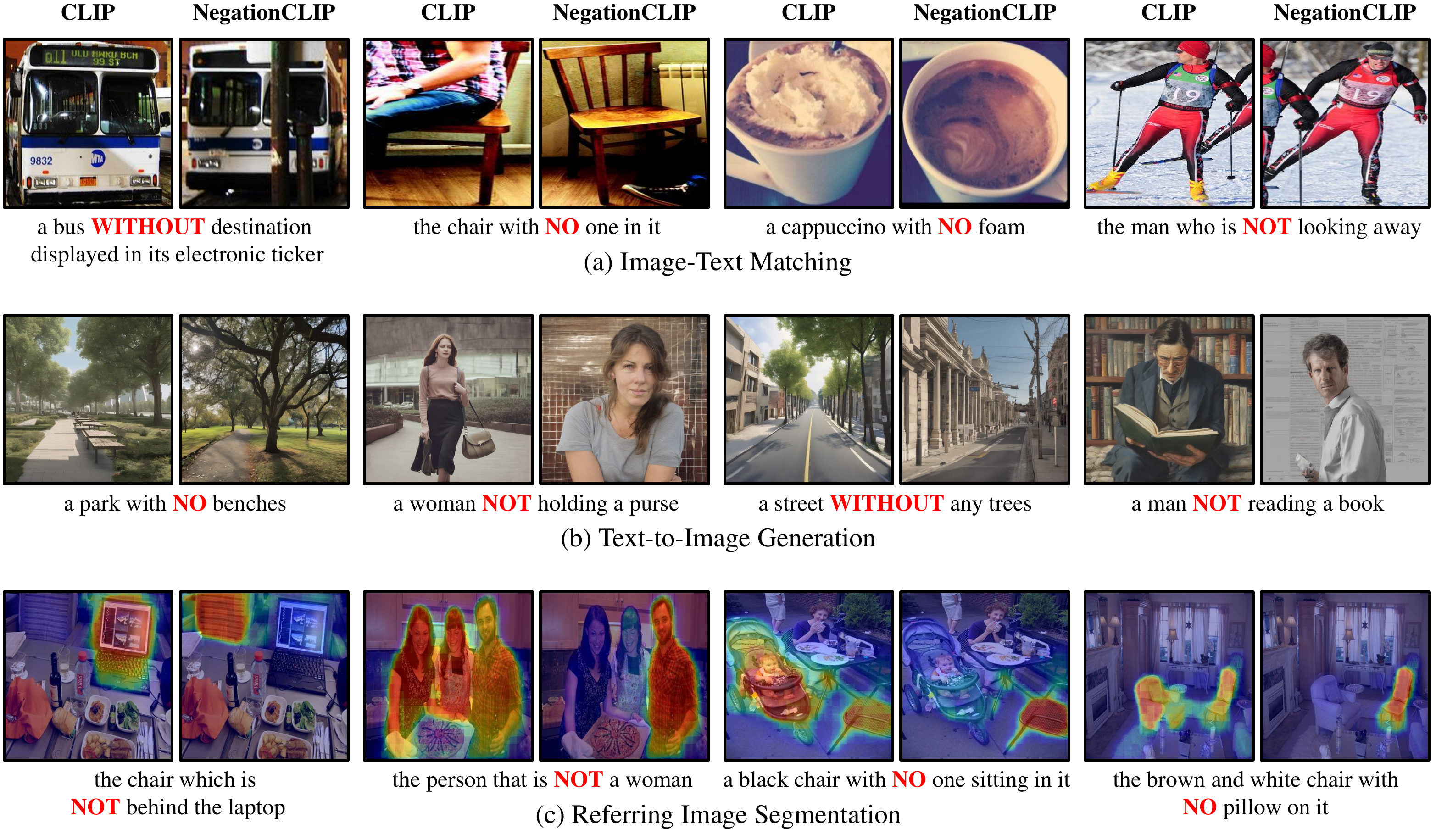}
    \captionof{figure}{Examples of the original CLIP and NegationCLIP (ours) on negation-inclusive data in multimodal tasks. Our NegationCLIP demonstrates a better understanding of negation concepts across various tasks.}
    \vspace{0.3em}
    \label{fig:main_figure}
\end{center}%
}]

\makeatletter
\renewcommand{\@makefntext}[1]{#1}
\makeatother
\footnotetext{\textsuperscript{†}Corresponding Authors}

\begin{abstract}
While CLIP has significantly advanced multimodal understanding by bridging vision and language, the inability to grasp negation---such as failing to differentiate concepts like ``parking'' from ``no parking''---poses substantial challenges.
By analyzing the data used in the public CLIP model's pre-training, we posit this limitation stems from a lack of negation-inclusive data.
To address this, we introduce data generation pipelines that employ a large language model (LLM) and a multimodal LLM to produce negation-inclusive captions.
Fine-tuning CLIP with data generated from our pipelines, we develop NegationCLIP, which enhances negation awareness while preserving the generality.
Moreover, to enable a comprehensive evaluation of negation understanding, we propose NegRefCOCOg—a benchmark tailored to test VLMs' ability to interpret negation across diverse expressions and positions within a sentence. 
Experiments on various CLIP architectures validate the effectiveness of our data generation pipelines in enhancing CLIP's ability to perceive negation accurately.
Additionally, NegationCLIP's enhanced negation awareness has practical applications across various multimodal tasks, demonstrated by performance gains in text-to-image generation and referring image segmentation.
\end{abstract}    
\section{Introduction}
\label{sec:intro}

Recent advances in vision-language models (VLMs)~\cite{albef,align,florence,blip,siglip} have demonstrated significant capabilities in integrating visual and linguistic information, achieving notable performance in multimodal tasks such as text-to-image (T2I) generation~\cite{flux2024, stablediffusion, p2} and referring image segmentation~\cite{clipseg, weaklyris, regionvlm}. 
Among these models, CLIP~\cite{clip} has emerged as particularly influential, serving as the foundation for numerous subsequent models~\cite{stablediffusion,llava,clipseg,dalle}. 
The effectiveness of these models, however, is inherently constrained by the capabilities of the CLIP encoder, underscoring the importance of its robustness. 

In response, research has sought to identify and address the inherent limitations of CLIP~\cite{cliplim1, alphaclip, fairerclip}, focusing on challenges related to the text encoder’s ability to understand sentence structure and relationships, which constrain compositional image-text alignment~\cite{whenandwhy, crepe, sugarcrepe}. 
Despite these efforts, the challenge of accurately handling negation remains largely underexplored, with only a few recent attempts to address this issue~\cite{clipbnl, conclip}.
Negation, marked by terms such as ``no,'' ``not,'' or ``without,'' plays a fundamental role in language by altering the meaning of words and phrases and, consequently, entire sentences. 
Therefore, a precise understanding of negation is crucial for CLIP to perform reliably.

In this study, our preliminary analysis (in ~\cref{sec:motivation}) reveals that CLIP frequently fails to capture the intended meaning of prompts involving negation. 
Experiments underscore this deficiency, demonstrating the need for targeted improvements in this area. 
Further investigation into the CLIP pre-training dataset~\cite{laion}  indicates that captions containing negation are underrepresented and, when present, often misaligned with the visual content, revealing a critical gap that impedes the model’s ability to understand negation.

To mitigate the limitation in data, we propose two data generation pipelines leveraging a large language model (LLM) and a multimodal LLM (MLLM) to create captions that incorporate negation and align accurately with images. 
The first pipeline generates negation terms based on the absence of contextually relevant objects, while the second pipeline expands the diversity of negation over object existence.
By fine-tuning the pre-trained CLIP text encoder with data generated from our proposed pipelines, we develop NegationCLIP, a model capable of improved comprehension of negation while maintaining general performance across various tasks.

Furthermore, evaluating negation comprehension in VLMs remains challenging due to limited benchmarks and exploration of this issue.
Although prior work~\cite{valse, crepe, conclip} has introduced image-to-text retrieval benchmarks, they still fall short of providing comprehensive coverage, often exhibiting bias against negation by classifying all negation-inclusive captions as incorrect or restricting negation to object existence.
In light of these limitations, we propose NegRefCOCOg, the first text-to-image retrieval benchmark for evaluating negation comprehension in VLMs.
NegRefCOCOg ensures that every retrieval query involves negation and supports a broader range of negation types, applying multiple negation terms not only to objects but also to attributes such as actions, adverbs, and prepositions.

Our experiments demonstrate that NegationCLIP, fine-tuned using data generated by our proposed framework, achieves superior negation understanding on both existing and newly developed NegRefCOCOg benchmarks while maintaining strong general task performance.
Additionally, NegationCLIP proves adaptable across various multimodal tasks.
Notably, replacing the text encoder in T2I generation models with that of NegationCLIP enhances negation comprehension—a capability often lacking in original T2I models (using the original CLIP encoder).
Furthermore, our NegationCLIP's text encoder enables improved performance in referring image segmentation for prompts containing negation, underscoring its scalability as a more contextually aware and flexible text encoder.
\cref{fig:main_figure} illustrates the versatility of NegationCLIP across diverse multimodal tasks involving negation. The key contributions of our study are summarized as follows:
\vspace{-3mm}

\paragraph{Contributions} 
1) We identify a significant limitation in CLIP’s ability to effectively process negation, tracing this issue to deficiencies in the pre-training dataset, where negation terms are underrepresented and poorly aligned with visual content.
2) We develop novel data generation pipelines that utilize a large language model (LLM) and a multimodal LLM (MLLM) to produce high-quality, negation-inclusive captions aligned with visual contexts, enhancing the training data for improved negation comprehension.
3) We propose NegRefCOCOg, a benchmark comprising various forms of negation, specifically designed to evaluate the negation comprehension capabilities of VLMs.
4) Our negation-aware model named NegationCLIP demonstrates robust negation comprehension and maintains generality, excelling across tasks such as image-text matching, T2I generation, and referring image segmentation.
\section{Negation: A Critical Challenge for CLIP}
\label{sec:motivation}

In this section, we design a simple experiment to demonstrate that CLIP struggles to handle negation effectively.
We then identify potential reasons behind this limitation from a data perspective.

\subsection{Case Study: Exposing the Negation Issue}
\label{subsec:celeba}

To evaluate CLIP’s ability to handle negation, we conduct a binary classification experiment using the CelebA~\cite{celeba} dataset, which contains 40 binary attributes for facial images.
For each of the 40 attributes, we construct positive and negative prompts following the CLIP-based image classification format: ``a photo of.''
For example, for the attribute ``eyeglasses,'' we generate prompts like ``a photo of a person wearing glasses'' and ``a photo of a person \textit{not} wearing glasses''.
Detailed prompts are provided in Appendix.

The classification task is structured as follows: given an image, CLIP is prompted with both the positive and negative prompts constructed above, and we evaluate the accuracy in matching the image with the correct prompt out of the two. 
We use balanced accuracy instead of standard accuracy to account for potential class imbalances between positive and negative examples. 

We report the average balanced accuracy of the CLIP ViT-L/14 model across all 40 attributes.
Despite this being a binary classification task---where random guessing yields an accuracy of 50\%---we obtained an average balanced accuracy of 60.8\%.
This performance is significantly low, considering that 1,000-way classification with ImageNet-1k produces 73.4\% accuracy using the same prompt format.
These findings highlight the need for targeted improvements to enhance CLIP's robustness in handling negation.

\subsection{Root Cause: Limited Negation in Training Data}
\label{subsec:laion}

We approach this issue from the perspective of pre-training data.
To investigate the presence and quality of negation, we analyze LAION-400M~\cite{laion}, the dataset popularly used for training public CLIP models such as OpenCLIP~\cite{openclip}.


\begin{table}[t]
    \centering
    \caption{Proportion of negation in captions and words within LAION-400M. The table shows the ratio of captions containing negation terms and the ratio of negation terms among all words.}
    {\footnotesize
    \begin{tabular}{lccc}
        \hline
        \textbf{Level} & \textbf{Total Count} & \textbf{Negation Count} & \textbf{Negation Ratio} \\
        \hline
        Caption & 414M & 2.91M & 0.70\% \\
        Word & 3.88B & 3.21M & 0.08\% \\  
        \hline
    \end{tabular}
    }
    \label{table:negation_ratio}
\end{table}

\begin{figure}[t]
    \centering
    \includegraphics[width=\columnwidth]{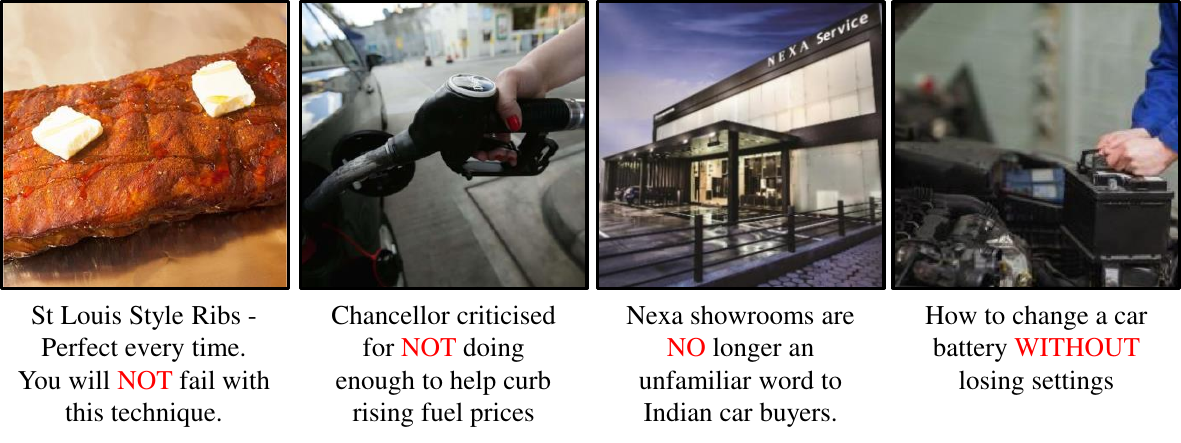}
    \caption{Examples of misleading negation samples in LAION-400M.}
    \vspace{-3mm}
    \label{fig:negation_misplacement}
\end{figure}

In \cref{table:negation_ratio}, we present the proportion of captions containing negation in LAION-400M. Our findings reveal that only about 0.704\% of captions in LAION-400M contain negation terms. Negation terms make up only 0.083\% of the total word count---an insufficient representation given the importance of negation in language. 
Furthermore, as illustrated in \cref{fig:negation_misplacement}, even when negation is present in captions, it often lacks alignment with the visual content of the image, providing no meaningful signal for the model to learn from. 
This scarcity of visually aligned negation can be attributed to the nature of image-text pairs typically used in VLM training.
Image-level captions naturally focus on describing the contents of an image, detailing what is visible, rather than specifying what is absent. 
As a result, the model receives minimal exposure to negation-related linguistic structures during training, making it difficult for CLIP to learn and respond to the exclusionary cues introduced by negation.

\begin{figure*}[!t]
  \includegraphics[width=\textwidth]{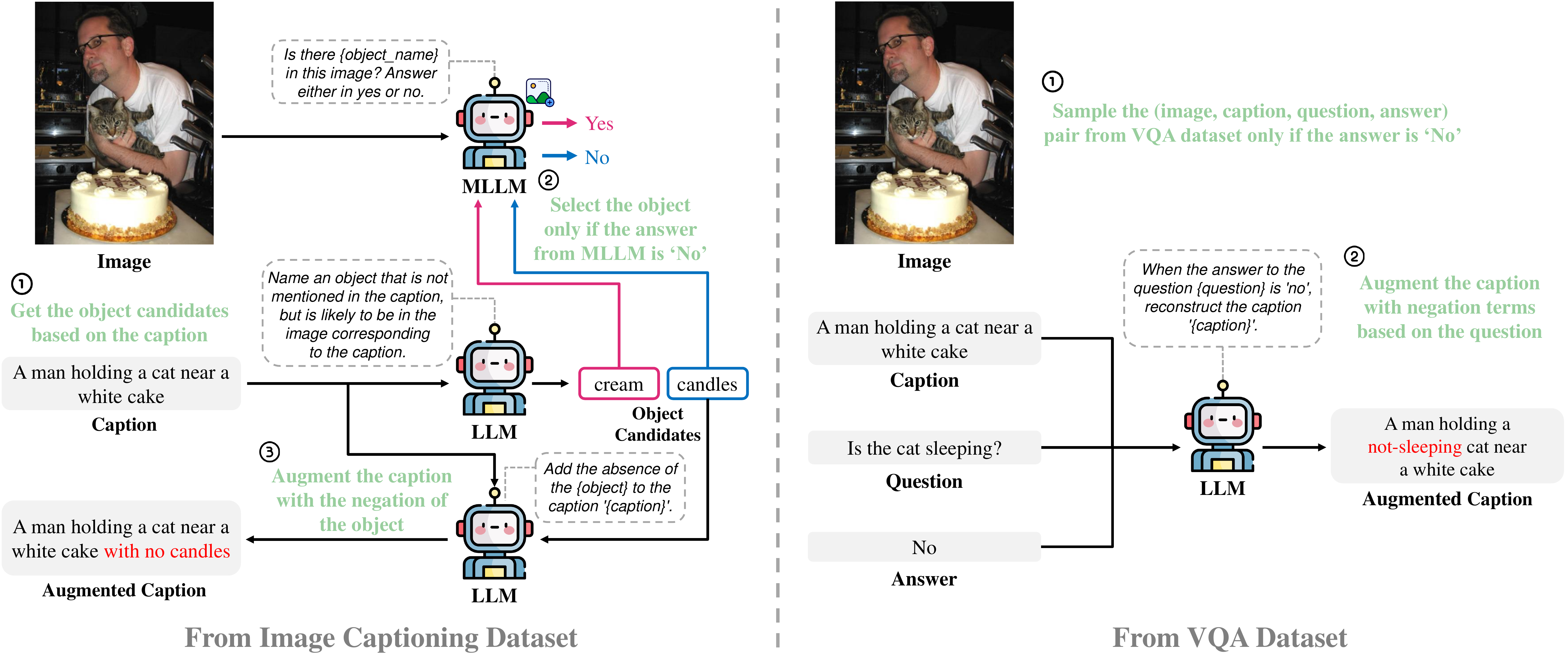}
  \caption{Data generation pipelines in our work. The left presents the pipeline of generating negation from object absence in the image captioning dataset, and the right presents the pipeline of generating negation from question-answer pairs in the VQA dataset.}
  \vspace{-3mm}
  \label{fig:pipeline}
\end{figure*}

These observations underline a critical need for diverse visually aligned negation-inclusive data, allowing the model to develop a more robust understanding of negation.
\section{Negation-Inclusive Data Generation}\label{sec:methods}

In this section, we introduce two data generation pipelines using LLM and MLLM to address the limitations in CLIP’s training data—the scarcity of negation terms and their misalignment with visual content.
After that, we propose NegRefCOCOg benchmark, which can provide an effective evaluation of negation awareness based on the existing referring image segmentation datasets.

\subsection{Generating Negation from Object Absence}
\label{subsec:pipeline1}

In this pipeline, we leverage an existing, well-established image captioning dataset, maximizing utility without the need for extensive new annotations.
The goal is to augment existing captions by naturally incorporating negation based on plausible objects that are likely to be present but are actually absent in the image. 
It has been demonstrated that the absence of objects in an image can be accurately determined by MLLMs~\cite{llava, instructblip, pope, mmhalbench}, making object-based negation a practical and reliable choice for our pipeline.

While the simplest approach in this pipeline might involve a negation regarding random objects, such a method is disconnected from the image context, making it less effective for supporting model learning.
Thus, we start by identifying plausible objects using an LLM, to which we provide the caption of the corresponding image.
We then use an MLLM to confirm the absence of these plausible objects in the image.
For objects confirmed to be absent, we augment the caption using the LLM to naturally incorporate negation terms with their absence, enriching the training data with contextually relevant negation examples. 
The overall process is as follows:

\begin{enumerate}
    \item \textbf{Extracting Plausible Object from Caption}: For each image-caption pair, we first provide only the caption to LLM to identify the plausible objects not mentioned in the caption but could reasonably be present in the image.
    \item \textbf{Verifying Object Absence with MLLM}:
    Since the process above does not consider the input image, there is a possibility that the plausible objects identified above may actually be present in the image.
    To confirm their absence, we provide the image to an MLLM and query it about the presence of the identified object. 
    This allows us to filter out objects that are present in the image.
    \item \textbf{Augmenting Caption with Negation}: For an object that the MLLM determines to be absent, we use an LLM to augment the original caption by adding information about the missing object with negation.
\end{enumerate}

\subsection{Expanding Diversity of Negation}
\label{subsec:pipeline2}
The pipeline above is effective but is limited to negations related to objects.
To further enrich the data while maximizing the utility of existing resources, we employ a secondary pipeline utilizing data sourced from the VQA dataset.
This pipeline introduces diversity in negation expressions by drawing on diverse question-answer pairs, specifically selecting pairs where answers are ``no.''
These pairs encompass not only object presence but also aspects such as actions being performed and attributes possessed by objects, allowing us to incorporate negation across a broader range of image content. 
This expansion provides the model with richer exposure to varied linguistic structures. 
The pipeline operates in the following steps:

\begin{enumerate}
    \item \textbf{Selecting ``No'' Data}: We identify image-question-answer triplets from the VQA dataset where the answer is ``no.'' 
    These pairs provide flexibility to incorporate various forms of negation, as they stem from a wide range of questions about different features within the image.
    \item \textbf{Augmenting Caption with Negation}: For each selected image and its original caption, we use LLM to augment the caption with negation terms based on the question and the corresponding answer. 
\end{enumerate}

The overall structure and process of the pipelines are illustrated in \cref{fig:pipeline}, and detailed prompts used for the pipeline are provided in Appendix.

\subsection{NegRefCOCOg Benchmark Proposal}\label{subsec:negrefcocog}

While efforts have been made to establish benchmarks for negation evaluation in VLMs, these benchmarks exhibit notable limitations.
CREPE Negate~\cite{crepe} and CC-Neg~\cite{conclip} assume that captions containing negation are always false, creating a shortcut where answers can be determined solely by the presence of negation.
This severe bias allows even a blind model to outperform VLMs in these benchmarks~\cite{sugarcrepe}.
VALSE~\cite{valse}, on the other hand, does not impose this bias but remains limited in scope, as it considers only a single negation term (``no'') and restricts negation to object existence.
In this work, we address these limitations by proposing NegRefCOCOg, a benchmark built upon the existing referring image segmentation datasets: RefCOCOg~\cite{refcoco}. 
Referring image segmentation datasets can be utilized as valuable sources for negation evaluation because their text prompts contain a relatively higher proportion of negation to distinguish between similar objects within the same image.
Additionally, they include diverse negation terms such as ``no,'' ``not,'' and ``without'' and extend negation to various positions within a sentence, covering actions and attributes in addition to object references.

In constructing NegRefCOCOg, we first sample prompts from RefCOCOg that include negation terms. 
Let \( T \) denote a negation-inclusive prompt.
For each \( T \), we identify a corresponding image patch \( P^{+} \), which aligns with \( T \), serving as the positive example. 
Additionally, we designate hard negative image patches \( P^{-} \), representing different instances of the same object category with \( P^{+} \) in a distinct location within the image. This approach ensures that \( P^{-} \) is of the same object type as \( P^{+} \) but does not align with the negated prompt.
After that, we filter and augment the image patches following the constraints that make our benchmark reliable and challenging for evaluating negation understanding.

\begin{figure}[t]
    \centering
    \includegraphics[width=\columnwidth]{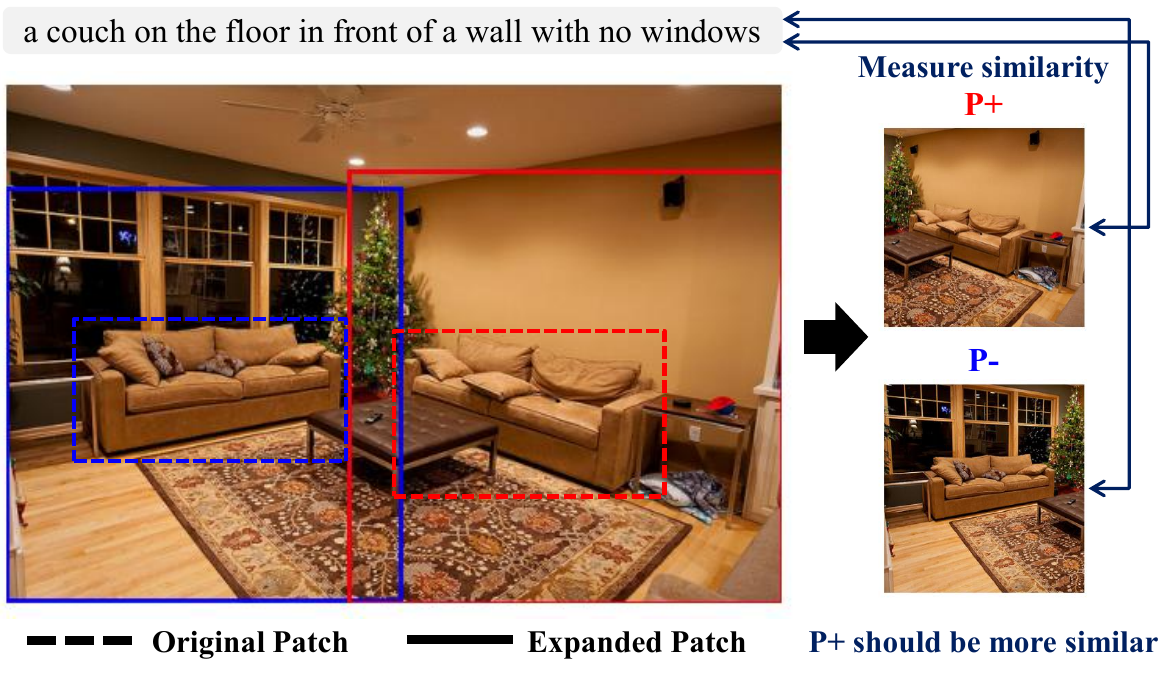}
    \caption{An example of NegRefCOCOg benchmark.}
    \label{fig:negrefcocog}
\end{figure}

For evaluation, the model calculates the similarity between the text embedding of \( T \) and the vision embeddings of \( P^{+} \) and \( P^{-} \).
If the similarity between \( T \) and \( P^{+} \) is greater than that between \( T \) and \( P^{-} \), the model scores 1; otherwise, it scores 0. 
\cref{fig:negrefcocog} presents an example of the NegRefCOCOg process. Detailed contents of the constraints and the evaluation can be found in Appendix.

\section{Experiments}
\label{sec:experiments}

We evaluate the effectiveness of our negation-inclusive data generation pipeline by fine-tuning the CLIP text encoder and comparing its performance against the original CLIP model and a negation baseline model. 
Our experiments aim to assess improvements in negation comprehension as well as general performance retention.

\subsection{Experimental Setup}

\paragraph{Experimental Details}

We generate a total of 229k image-text pairs using our data generation pipelines, with 147k pairs from the first pipeline (\cref{subsec:pipeline1}) and 82k pairs from the second (\cref{subsec:pipeline2}).
For the pipelines, we employ Llama-3-8B~\cite{llama3} as the LLM and LLaVA-1.6~\cite{llava1.6} as the MLLM.
For datasets, we use COCO~\cite{coco} as the image captioning dataset and VQAv2~\cite{vqav2} for the VQA dataset.

For fine-tuning, we freeze the vision encoder and fine-tune only the text encoder. This approach helps preserve the original embedding space, making our model, NegationCLIP, adaptable across various tasks without requiring further adjustments.
We use the standard InfoNCE loss~\cite{infonce} with a learning rate of $1\mathrm{e}{-6}$, optimized with the AdamW~\cite{adamw} optimizer. Additional prompt design details and specific configurations are provided in Appendix.

\paragraph{Models}

We evaluate our model, NegationCLIP, with three baselines:

\begin{itemize}[leftmargin=*, nolistsep]
    \item{
    CLIP~\cite{clip}: The original, pre-trained CLIP model without any modifications.}
    \item{
    CLIP-bnl~\cite{clipbnl}, CoN-CLIP~\cite{conclip}: Baseline models that have been fine-tuned specifically for negation comprehension, which provides a direct comparison for assessing our approach.}
    \item{
    NegationCLIP: Our proposed model, fine-tuned on negation-inclusive data generated by our two pipelines.
    }
\end{itemize}

\paragraph{Benchmarks}
We employ the following benchmarks to assess each model’s performance comprehensively. 

\begin{itemize}[leftmargin=*, nolistsep]
    \item{
    NegRefCOCOg: Our primary benchmark for assessing VLMs' negation comprehension in an image-text matching task. The model is given a negation-inclusive prompt and two image candidates, and the correct match of prompt and image must be selected.
    }
    \item{
    VALSE~\cite{valse} Existence: This benchmark evaluates VLMs' ability to handle negation in an image-to-text retrieval task. Given an image and two prompts—one indicating an object’s presence and the other its absence—the model must select the prompt that aligns with the image.
    }
    \item{
    ImageNet~\cite{imagenet} classification \& COCO~\cite{coco} retrieval: We evaluate the model’s zero-shot classification accuracy on ImageNet and its recall@5 on COCO text-to-image retrieval, confirming that fine-tuning on negation-inclusive data does not degrade its general capability.
    } 
\end{itemize}

\subsection{Main Results}

\begin{table}[t]
    \centering
    \caption{Comparison of model performance on negation and general benchmarks across different architectures. 
    We include all publicly available CLIP-bnl and CoN-CLIP checkpoints in addition to the baseline CLIP and NegationCLIP (Ours).
    }
    \resizebox{1.0\linewidth}{!}{
    \begin{tabular}{lccccc}
        \toprule
        \multirow{2}{*}{\textbf{Model}} & \multirow{2}{*}{\textbf{Arch.}} & \multicolumn{2}{c}{\textbf{Negation Benchmarks}} & \multicolumn{2}{c}{\textbf{General Benchmarks}} \\
        \cmidrule(lr){3-4} \cmidrule(lr){5-6}
                       &                       & \textbf{VALSE} & \textbf{NegRefCOCOg} & \textbf{ImageNet} & \textbf{COCO} \\
        \midrule
        CLIP & \multirow{4}{*}{ViT-B/32}   & 70.97 & 57.73 & 62.02 & 54.78 \\
        CLIP-bnl~\cite{clipbnl} &          & 76.78 & 62.05 & 53.33 & 55.47 \\
        CoN-CLIP~\cite{conclip} & & 71.72 & 55.45 & \textbf{63.08} & 55.66 \\
        NegationCLIP &                     & \textbf{80.15} & \textbf{64.09} & 60.97 & \textbf{68.00} \\
        \midrule
        CLIP & \multirow{3}{*}{ViT-B/16}   & 69.48 & 58.64 & 66.71 & 57.75 \\
        CoN-CLIP~\cite{conclip} & & 62.55 & 55.68 & \textbf{68.57} & 59.83 \\
        NegationCLIP &                     & \textbf{80.52} & \textbf{64.32} & 66.33 & \textbf{70.57} \\
        \midrule
        CLIP & \multirow{3}{*}{ViT-L/14}   & 66.85 & 57.27 & 73.44 & 59.99 \\
        CoN-CLIP~\cite{conclip} & & 65.73 & 55.45 & \textbf{75.38} & 63.18 \\
        NegationCLIP &                     & \textbf{79.59} & \textbf{62.95} & 73.91 & \textbf{72.77} \\
        \midrule
        CLIP & \multirow{2}{*}{\shortstack{ViT-L/14 \\ @336px}} & 64.61 & 57.05 & 74.92 & 60.74 \\
        NegationCLIP &                 & \textbf{78.65} & \textbf{62.95} & \textbf{75.15} & \textbf{73.43} \\
        \bottomrule
    \end{tabular}
    }
    \label{table:architecture_results}
\end{table}

\cref{table:architecture_results} summarizes the performance of each model across the three evaluation benchmarks.

For \textbf{ImageNet} classification and \textbf{COCO} retrieval, NegationCLIP maintains accuracy comparable to, or even exceeding, the original CLIP, particularly with larger architectures.  
This demonstrates that our fine-tuning approach for negation comprehension does not compromise the model’s general capabilities.  

For \textbf{NegRefCOCOg} and \textbf{VALSE} Existence, NegationCLIP consistently outperforms the original CLIP and all baseline models across all architectures, demonstrating its effectiveness in interpreting negations.  
This strong performance can be attributed to how NegationCLIP integrates negation-inclusive captions during training.  
Unlike CoN-CLIP, which first generates negated captions and then retrieves semantically similar images, our approach begins with an image and generates a corresponding caption that accurately captures its negation-related attributes.  
By grounding negation-inclusive captions directly in image content, our model learns to better capture the semantic relationships of negated concepts, leading to superior performance in benchmarks requiring fine-grained negation understanding.  

Overall, these results confirm that the negation-inclusive data generation pipeline not only significantly improves negation comprehension but also preserves general performance, making NegationCLIP a robust model for handling both negation-specific and broader vision-language tasks.

\subsection{Ablation Study on Data Configurations}

\begin{table}[t]
    \centering
    \caption{Ablation results on different data configurations.}
    {\footnotesize
    \begin{tabular}{llcc} 
        \toprule
        \textbf{Arch.} & \textbf{Data Config.} & \textbf{VALSE} ↑ & \textbf{NegRefCOCOg} ↑ \\
        \midrule
        \multirow{5}{*}{ViT-B/32}     & Original            & 70.97 & 57.73 \\
                                      & + Rand-P1           & 73.78 & 62.05 \\
        \cmidrule(lr){2-4}
                                         & + P1                & \textbf{80.15} & 63.18 \\
                                      & + P2                & 76.78 & \textbf{64.32} \\
                                      & + P1 + P2            & \textbf{80.15} & 64.09 \\
        \midrule
        \multirow{5}{*}{ViT-B/16}     & Original             & 69.48 & 58.64 \\
                                      & + Rand-P1            & 76.22 & 60.91 \\
        \cmidrule(lr){2-4} 
                                      & + P1                 & 77.53 & 63.41 \\
                                      & + P2                 & 80.15 & 63.41 \\
                                      & + P1 + P2            & \textbf{80.52} & \textbf{64.32} \\
        \midrule
        \multirow{5}{*}{ViT-L/14}     & Original     & 66.85 & 57.27 \\
                                      & + Rand-P1            & 76.40 & 60.23 \\
        \cmidrule(lr){2-4}
                                      & + P1                 & 77.53 & 60.23 \\
                                      & + P2                 & 76.03 & 62.27 \\
                                      & + P1 + P2            & \textbf{79.59} & \textbf{62.95} \\
        \midrule
        \multirow{5}{*}{\shortstack{ViT-L/14 \\ @336px}} & Original  & 64.61 & 57.05 \\
                                      & + Rand-P1           & 76.22 & 60.00 \\
        \cmidrule(lr){2-4}
                                      & + P1                 & 79.78 & 60.91 \\
                                      & + P2                 & 75.28 & 61.59 \\
                                      & + P1 + P2            & \textbf{80.34} & \textbf{62.95} \\
        \bottomrule
    \end{tabular}
    }
    \label{table:ablation}
\end{table}

In this section, we analyze the impact of different data configurations used for fine-tuning to evaluate how effectively they improve the model's handling of negation. P1 and P2 denote our proposed pipelines described in \cref{subsec:pipeline1} and \cref{subsec:pipeline2} respectively. 
Additionally, we include a Rand-P1 configuration, similar to P1 but using randomly selected objects instead of plausible ones. For more ablation study, please refer to Appendix.

As shown in \cref{table:ablation}, across all architectures, the Rand-P1 configuration consistently underperforms compared to P1. This result underscores the importance of plausible object selection in P1, as random object choices lack context, making them less effective for training.

The results indicate that combining P1 and P2 yields the best performance across most negation benchmarks, suggesting that both object-based and VQA-derived negation contribute complementary benefits. While P1 alone performs competitively on the VALSE benchmark, adding P2 further improves performance, especially on NegRefCOCOg. This trend suggests that NegRefCOCOg, in contrast to VALSE, better captures the diversity of negation expressions, including those based on actions and attributes. Consequently, the combined configuration (P1 + P2) aligns well with NegRefCOCOg’s expanded scope, leading to superior performance in negation understanding.
\section{Application}
\label{sec:application}

To further validate our negation-inclusive fine-tuning approach, we apply the model across different multimodal tasks. Specifically, we evaluate its performance on T2I generation and referring image segmentation, demonstrating improvements in negation comprehension.

\subsection{Text-to-Image Generation with Negation}
\label{subsec:t2i}

\begin{figure*}[t]
    \centering
    \includegraphics[width=1.\linewidth]{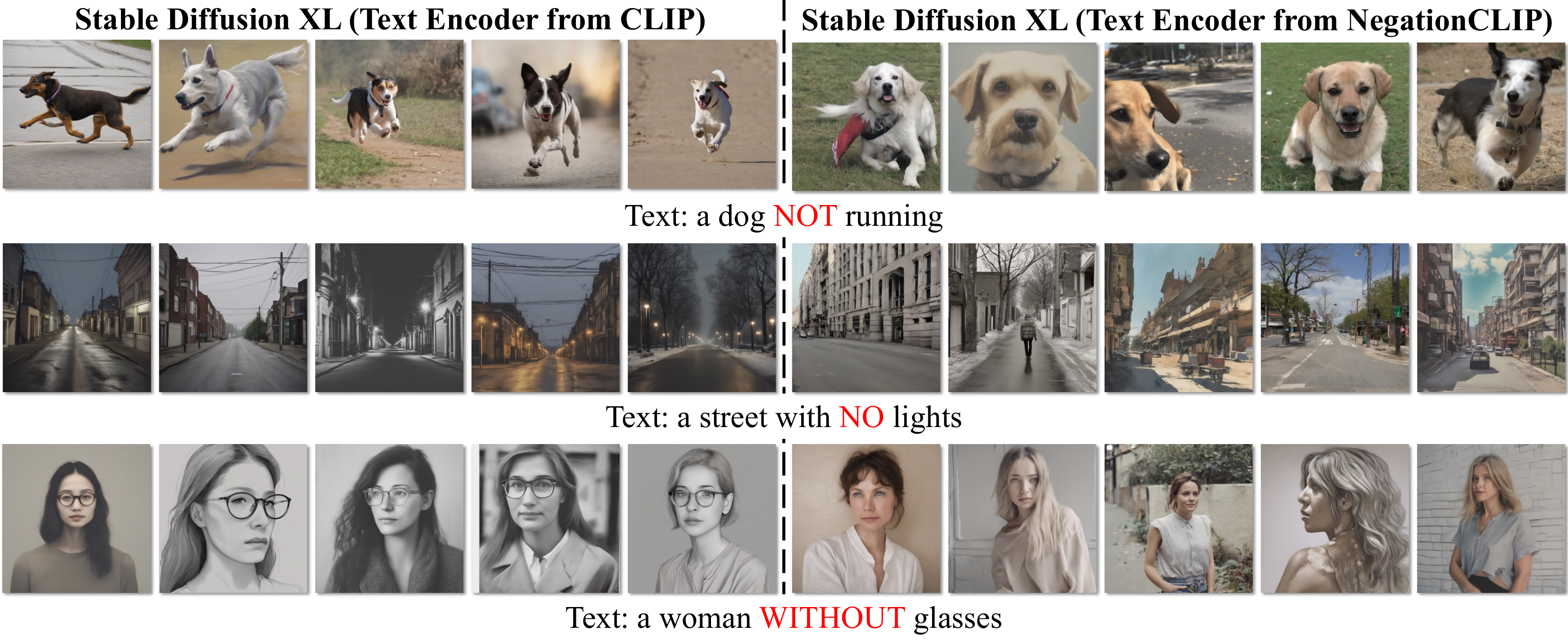}
    \caption{Examples of text-to-image generation on negation-inclusive prompts.}
    \label{fig:t2i_results}
\end{figure*}

T2I models often rely on the CLIP text encoder for interpreting prompts but struggle with negation, a limitation that aligns with CLIP’s challenges in handling negation. 
For instance, given a prompt like ``a man not wearing a hat,'' a T2I model using the original CLIP text encoder may generate an image of a man wearing a hat, ignoring the negation.

To evaluate our model’s ability to handle negation in T2I tasks, we replace the original CLIP text encoders in Stable Diffusion~\cite{stablediffusion} models with our NegationCLIP text encoder, without further training on the T2I model. 
This direct substitution is possible because we fine-tune only the text encoder, preserving the original image embedding space and maintaining alignment with it.

We utilize ChatGPT~\cite{chatgpt} to generate a set of 107 negation-inclusive prompts to measure each model's ability to represent negated concepts in generated images.
We assess the performance of negation comprehension using an MLLM evaluator, mPLUG~\cite{mplug}, inspired by the evaluation methodology employed in TIFA~\cite{tifa}.
We provide the MLLM with the generated image and two queries:  
(1) whether the subject is present in the image, and  
(2) whether the negation-related concept is correctly excluded.  
If both conditions are met, we assign a score of 1; otherwise, a score of 0.  
For details on the specific prompts used for image generation and MLLM evaluation, please refer to Appendix.

\begin{table}[t]
    \centering
    \caption{Comparison of text-to-image generation performance on negation-inclusive prompts.}
    {\footnotesize
    \begin{tabular}{lcc}
        \toprule
        \textbf{Model} & \textbf{TIFA \(\uparrow\)} & \textbf{Neg Score \(\uparrow\)} \\
        \midrule
        SD-1.4~\cite{sd1.4} & 0.786 & 0.295 \\
        SD-1.4 w/ CoN-CLIP text encoder & 0.783 & 0.243 \\
        SD-1.4 w/ NegationCLIP text encoder & \textbf{0.790} & \textbf{0.449} \\
        \cmidrule(lr){1-3}
        SDXL-1.0~\cite{sdxl} & \textbf{0.849} & 0.308 \\
        SDXL-1.0 w/ NegationCLIP text encoder & 0.802 & \textbf{0.421} \\
        \bottomrule
    \end{tabular}%
    }
    \label{table:t2i_results}
\end{table}

For each of the 107 prompts, we generate images using 5 different random seeds, and the average score across these generations is reported as the \textbf{Neg Score} in \cref{table:t2i_results}.
Our model achieves significantly higher Neg Scores across both SD-1.4~\cite{sd1.4} and SDXL-1.0~\cite{sdxl}, with over 0.15 improvement compared to the original text encoder, indicating enhanced negation comprehension, while CoN-CLIP’s low Neg Score further confirms the issues arising from the misalignment between its negation-inclusive captions and retrieved images, as observed in retrieval benchmarks.
Qualitatively, as shown in \cref{fig:t2i_results}, our model successfully generates images reflecting prompts like ``a dog not running'' or ``a street with no lights,'' whereas the original model often fails to capture the negation.

To ensure that negation-aware fine-tuning of CLIP does not compromise general T2I quality, we also evaluate each model on the TIFA~\cite{tifa} benchmark, which assesses general text-image alignment. 
\cref{table:t2i_results} shows that our model retains competitive TIFA scores, slightly exceeding the original SD-1.4 model and trailing slightly in SDXL-1.0.
The slight decrease in SDXL’s performance may stem from its use of dual text encoders, both of which were replaced with our negation-aware encoder.
This setup could introduce minor alignment challenges, which might be further improved through the additional adaption of diffusion models to our NegationCLIP text encoder.

Overall, our approach addresses a key limitation in NegationCLIP-based T2I models, enabling more accurate generation in negation contexts.

\begin{table}[t]
    \centering
    \caption{Comparison of referring image segmentation performance on PhraseCut and RefCOCOg (Neg). }
    {\footnotesize
    \begin{tabular}{lcccc}
        \toprule
        \multirow{2}{*}{\textbf{Model}} & \multicolumn{2}{c}{\textbf{PhraseCut}} & \multicolumn{2}{c}{\textbf{RefCOCOg (Neg)}} \\
        \cmidrule(lr){2-3} \cmidrule(lr){4-5}
                       & \textbf{mIoU} & \textbf{IoU\textsubscript{BIN}} & \textbf{mIoU} & \textbf{IoU\textsubscript{BIN}} \\
        \midrule
        CLIPSeg~\cite{clipseg} & \textbf{0.562} & 0.736 & 0.267 & 0.492 \\
        CoN-CLIPSeg & 0.539 & 0.724  & 0.123 & 0.379 \\
        NegationCLIPSeg & 0.561 & \textbf{0.737} & \textbf{0.288} & \textbf{0.521} \\
        \bottomrule
    \end{tabular}%
    }
    \label{table:segmentation_results}
\end{table}

\subsection{Referring Image Segmentation}

Referring image segmentation task requires models to segment regions within an image corresponding to the given text prompts. 
To assess the effectiveness of our NegationCLIP, we replace the text encoder in the existing referring image segmentation model, CLIPSeg~\cite{clipseg}, with NegationCLIP text encoder. We simply replace the text encoder without further training as we did in \cref{subsec:t2i}. We refer to this model as NegationCLIPSeg and the same goes for CoN-CLIPSeg.

\cref{table:segmentation_results} presents the mIoU and IOU\textsubscript{BIN} as done in CLIPSeg~\cite{clipseg}.
On the PhraseCut dataset, which lacks negation in prompts, the performance of NegationCLIPSeg is comparable to the original model. On the negation-inclusive subset of RefCOCOg, NegationCLIPSeg demonstrates a performance boost, achieving higher mIoU and IoU\textsubscript{BIN} scores compared to the original CLIPSeg. This improvement highlights NegationCLIPSeg's enhanced capability to handle prompts with negation, while still retaining general performance.

\begin{figure}[t]
    \centering
    \includegraphics[width=\linewidth]{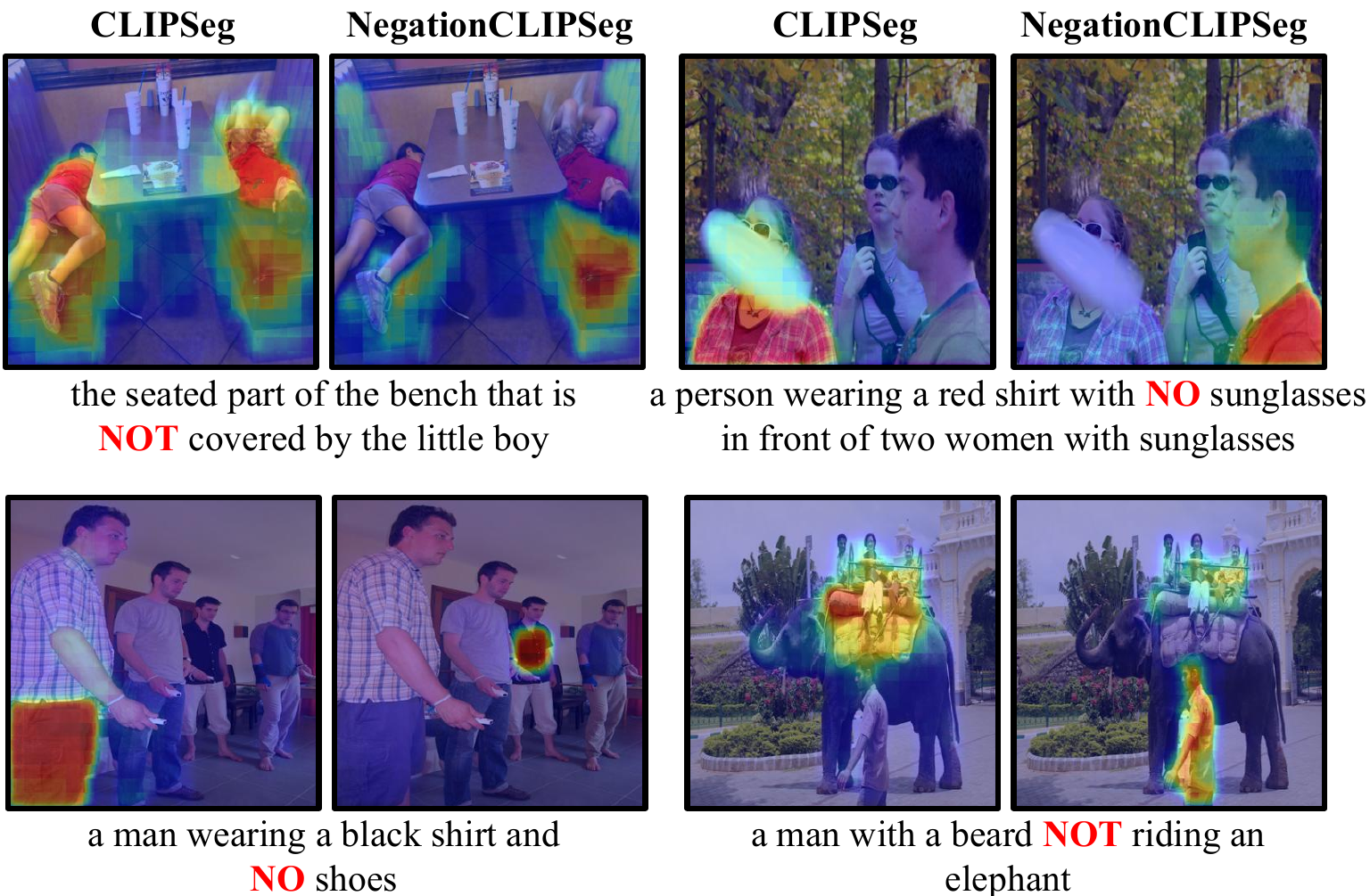}
    \vspace{-1em}
    \caption{Examples of referring image segmentation on negation-inclusive prompts.}
    \vspace{-1em}
    \label{fig:ris_results}
\end{figure}

\cref{fig:ris_results} shows qualitative examples from the RefCOCOg negation subset. In each example, NegationCLIPSeg generates more accurate segmentations that align with the negation-specific prompts, whereas the original CLIPSeg model often fails to correctly interpret the negation. For instance, given the prompt “a man with a beard \textit{not} riding an elephant,” NegationCLIPSeg successfully excludes the people riding an elephant, focusing mostly on the man not riding an elephant. 

Overall, these results highlight the potential of incorporating a negation-aware text encoder to improve performance in referring image segmentation tasks in scenarios that require precise comprehension of negation within multimodal contexts.

\section{Related Work}
\label{sec:related}

\paragraph{Vision-Language Models and Their Limitations}
VLMs~\cite{clip, align, florence} learn joint image-text embeddings from large-scale paired data, enabling diverse multimodal tasks such as image classification and zero-shot retrieval.

However, numerous studies have highlighted limitations in VLMs, particularly concerning the models’ handling of compositional language, object relationships, and fine-grained language distinctions~\cite{whenandwhy, crepe, sugarcrepe}. 
To address such limitations, researchers have explored targeted data generation techniques and fine-tuning methods. 
Data augmentation approaches~\cite{allseeing, picture77} aim to improve the robustness of VLMs by generating or augmenting data with challenging linguistic structures.
Despite these improvements, the issue of negation remains relatively underexplored.

\vspace{-3mm}
\paragraph{Understanding and Addressing Negation}
Negation has been a challenging aspect in language models, as it plays a crucial role in altering or reversing the meaning of a phrase.
While some research has addressed negation comprehension in natural language processing (NLP), such as studies highlighting significant issues in BERT’s~\cite{bert} interpretation of negated statements~\cite{negbert, notadataset, understandingnot}, there has been relatively little exploration of negation within multimodal models. 
In the vision-language domain, only a handful of studies have specifically focused on incorporating negation, and even these efforts have limited scope. 

Existing studies have integrated negation into CLIP for out-of-distribution (OoD) detection tasks~\cite{clipn, oodclip}, while others have focused on specific applications, such as negation handling in video tasks~\cite{clipbnl}.
However, these models are restricted to the specific contexts, making these approaches less adaptable for broader downstream tasks.
A more recent approach~\cite{conclip} seeks to enhance CLIP’s ability to process negation beyond task-specific settings  through fine-tuning, using image-caption pairs where negation-inclusive captions are generated first and then matched with similar images retrieved from a fixed pool.

Our work extends this line of research by developing a negation-inclusive text encoder within CLIP, designed for flexible, plug-and-play integration across a variety of multimodal tasks, addressing the underexplored challenge of negation comprehension in VLMs.
\section{Conclusion}
\label{sec:conclusion}

This study addresses a critical limitation in CLIP by introducing a negation-inclusive fine-tuning approach that significantly enhances the models’ ability to interpret negated prompts. Our data generation pipeline leverages large language models to create diverse negation-inclusive captions, enabling fine-tuning that effectively bridges the gap in negation comprehension. Experimental results on the VALSE benchmark and our proposed NegRefCOCOg benchmark demonstrate substantial improvements in our NegationCLIP over the original CLIP in handling negation, while maintaining strong performance on general benchmarks. Furthermore, we show that NegationCLIP enables effective applications of negation across various multimodal tasks, including text-to-image generation and referring image segmentation.

Our work underscores the potential of targeted data generation in advancing the semantic capabilities of CLIP. By refining their ability to process negation, we move closer to developing a VLM that is more aligned with the subtleties of human language, ultimately making them better suited for complex multimodal tasks.

\section*{Acknowledgements}
This work was supported by the National Research Foundation of Korea (NRF) grant funded by the Korea government (MSIT) (No. 2022R1A3B1077720, 2022R1A5A7083908, RS-2025-00555943), Institute of Information \& Communications Technology Planning \& Evaluation (IITP) grant funded by the Korea government (MSIT) [NO. RS-2021-II211343, RS-2022-II220959, Artificial Intelligence Graduate School Program (Seoul National University), No.RS-2025-02263754, Human-Centric Embodied AI Agents with Autonomous Decision-Making], the BK21 FOUR program of the Education and Research Program for Future ICT Pioneers, Seoul National University in 2025, and Samsung Electronics (IO221213-04119-01).

{
    \small
    \bibliographystyle{ieeenat_fullname}
    \bibliography{main}
}

\clearpage
\maketitlesupplementary

\appendix
\section{OpenCLIP Training Dataset Statistics}
\label{sec:data_statistics}

In this section, we provide additional statistics on negation-related terms in the datasets used for OpenCLIP~\cite{openclip} training, complementing the analysis presented in Sec. 2.2. 
Specifically, we report the frequency of the negation terms ``no,'' ``not,'' and ``without'' in the DataComp-1B~\cite{datacomp} and LAION-2B~\cite{laion} datasets in \cref{table:datacomp_laion}. 
Both datasets exhibit trends similar to those observed in LAION-400M~\cite{laion}.
These findings are consistent with the statistics we reported for LAION-400M, highlighting the insufficient representation of negation in datasets used for OpenCLIP pre-training.

\begin{table}[ht]
    \centering
    \caption{Proportion of negation in captions and words within DataComp-1B and LAION-2B.}
    \resizebox{1.\columnwidth}{!}
    {%
    \begin{tabular}{llccc}
        \toprule
         \textbf{Dataset}& \textbf{Level} & \textbf{Total Count} & \textbf{Neg. Count} & \textbf{Neg. Ratio} \\
        \midrule
        \multirow{2}{*}{DataComp-1B~\cite{datacomp}} & Caption & 1.38B & 10.4M & 0.75\% \\
        & Word & 13.8B & 11.8M & 0.09\% \\  
        \midrule
        \multirow{2}{*}{LAION-2B~\cite{laion}} & Caption & 2.08B & 19.3M & 0.93\% \\
        & Word & 21.9B & 21.1M & 0.10\% \\  
        \bottomrule
    \end{tabular}
    }
    \label{table:datacomp_laion}
\end{table}
\section{More Ablation Studies}
\label{sec:ablation}

In addition to the results presented in Sec. 4.3, we further evaluate the impact of using original captions instead of the generated captions for fine-tuning on the same set of images. 

The results shown in \cref{table:further_ablation} demonstrate that incorporating our generated captions consistently achieves the highest performance on both the VALSE and NegRefCOCOg benchmarks, across all architectures.
Notably, while fine-tuning with original captions can also lead to degradation, as observed in the ViT-B/32 architecture on the VALSE benchmark, fine-tuning with our generated captions consistently improves performance. 
This underscores the efficacy of our data generation pipeline in enhancing negation comprehension.

\begin{table}[t]
    \centering
    \caption{More ablation results on different data configurations.}
    {\footnotesize
    \begin{tabular}{llcc} 
        \toprule
        \textbf{Arch.} & \textbf{Data Config.} & \textbf{VALSE} ↑ & \textbf{NegRefCOCOg} ↑ \\
        \midrule
        \multirow{3}{*}{ViT-B/32}     & Original & 70.97 & 57.73 \\
                                      & + Original Caption & 68.35 & 60.45 \\
        \cmidrule{2-4}
                                      & + Our Caption & \textbf{80.15} & \textbf{64.09} \\
        \midrule
        \multirow{3}{*}{ViT-B/16}     & Original & 69.48 & 58.64 \\
                                      & + Original Caption & 73.97 & 60.91 \\
        \cmidrule{2-4}
                                      & + Our Caption & \textbf{80.52} & \textbf{64.32} \\
        \midrule
        \multirow{3}{*}{ViT-L/14}     & Original & 66.85 & 57.27 \\
                                      & + Original Caption & 74.91 & 60.00 \\
        \cmidrule{2-4}
                                      & + Our Caption & \textbf{79.59} & \textbf{62.95} \\
        \midrule
        \multirow{3}{*}{\shortstack{ViT-L/14 \\ @336px}} & Original & 64.61 & 57.05 \\
                                      & + Original Caption & 73.97 & 58.41 \\
        \cmidrule{2-4}
                                      & + Our Caption & \textbf{80.34} & \textbf{62.95} \\
        \bottomrule
    \end{tabular}
    }
    \label{table:further_ablation}
\end{table}

\section{Data Generation Pipelines}
\label{sec:datagen_pipelines}

We provide additional details on the two data generation pipelines discussed in Sec. 3.1 and Sec. 3.2 of the main paper. 
Specifically, we include the prompts used with the LLM and MLLM during data generation and provide examples of the generated image-caption pairs.
The prompts used in the two pipelines are presented in \cref{table:prompt_pipeline1} and \cref{table:prompt_pipeline2} respectively. 

\begin{figure*}[t]
    \centering
    \includegraphics[width=1.\linewidth]{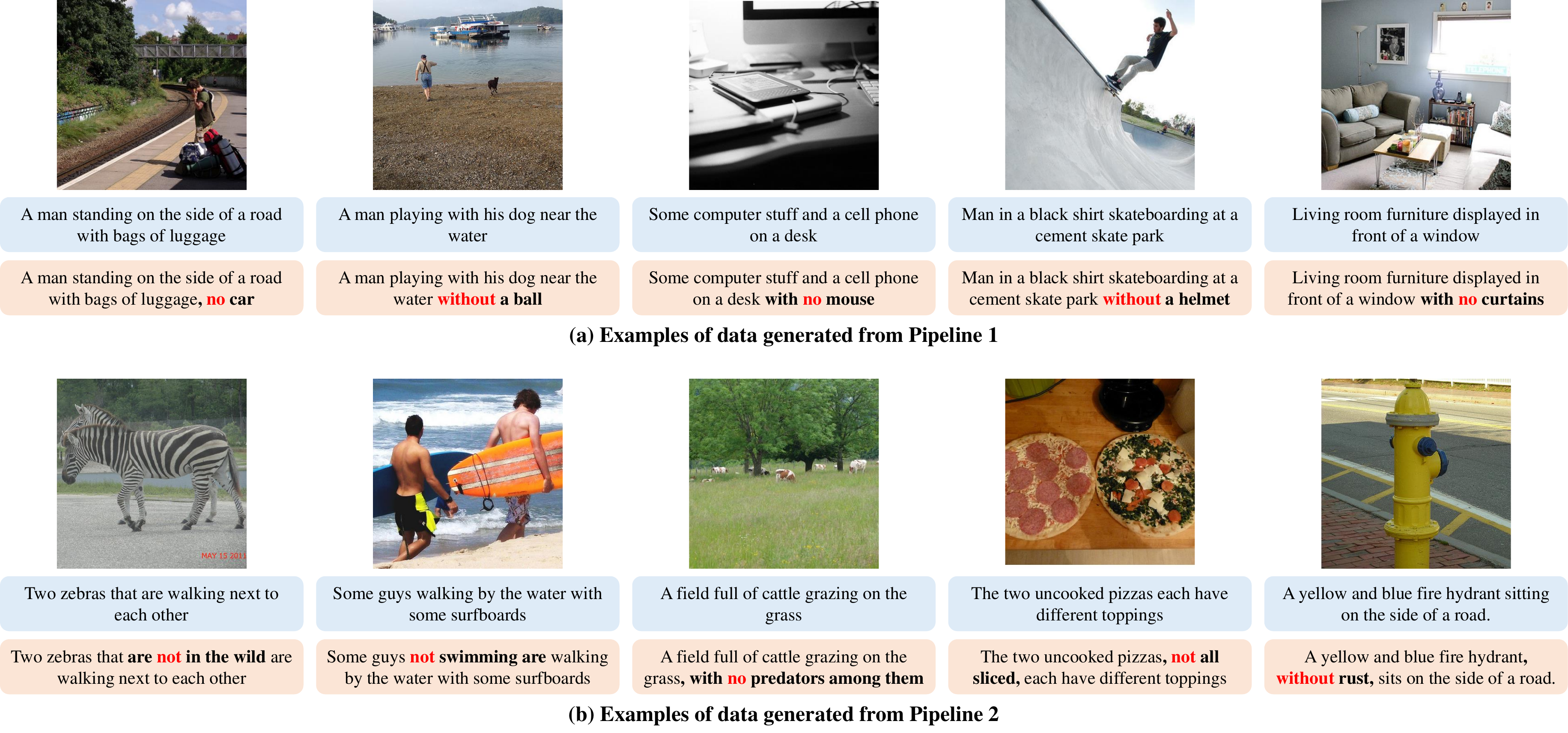}
    \caption{Examples of data generated by our proposed pipelines. (a) demonstrates captions augmented through Pipeline 1, and (b) illustrates captions augmented through Pipeline 2. \textbf{Blue boxes} represent the original captions, and \textbf{red boxes} show the augmented captions with negation terms.}
    \label{fig:datagen_examples}
\end{figure*}

We provide qualitative examples of the data generated by our proposed data generation pipelines in \cref{fig:datagen_examples}. 
As shown in \cref{fig:datagen_examples} (a), captions generated by Pipeline 1 accurately incorporate the absence of objects such as ``car,'' ``ball,'' or ``curtains,'' which are contextually plausible within the scene. 
This process enhances the training data with negation terms while maintaining alignment with the image content.
\cref{fig:datagen_examples} (b) demonstrates how Pipeline 2 captures a broader scope of negation, such as negating actions (e.g., ``not swimming''), adjectival phrases (e.g., ``not in the wild'').
These examples highlight the flexibility and effectiveness of this pipeline in generating rich negation-inclusive captions.

\begin{table}[t]
\centering
\caption{Prompts used in our Pipeline 1.}
{\footnotesize
\renewcommand{\arraystretch}{1.2}
\begin{tabular}{p{0.1\columnwidth}|p{0.8\columnwidth}}
\midrule
\multicolumn{2}{c}{\textbf{Step 1: Extracting Plausible Object from Caption}} \\
\midrule
\textbf{System} & \texttt{You are a helpful chatbot that answers with only one word.} \\
\hline
\textbf{User} & \texttt{Name an object that is not mentioned in the caption, but is likely to be in the image corresponding to the caption '\{caption\}'.} \\
\hline
\textbf{LLM} & \texttt{\{object\}.} \\
\midrule

\multicolumn{2}{c}{\textbf{Step 2: Verifying Object Absence with MLLM}} \\
\midrule
\textbf{System} & \texttt{A chat between a curious human and an artificial intelligence assistant. The assistant gives helpful, detailed, and polite answers to the human's questions.} \\
\hline

\textbf{User} & \texttt{<image>} \newline 
\texttt{Is there \{object\} in this image? Answer either yes or no.} \\
\hline
\textbf{MLLM} & \texttt{\{yes/no\}.} \\
\midrule

\multicolumn{2}{c}{\textbf{Step 3: Augmenting Caption with Negation}} \\
\midrule
\textbf{System} & \texttt{You are a helpful chatbot that generates concise caption.} \\
\hline
\textbf{User} & \texttt{Add the absence of the \{object\} to the caption '\{caption\}'.} \\
\hline
\textbf{LLM} & \texttt{\{updated\_caption\}.} \\
\midrule
\end{tabular}
}
\label{table:prompt_pipeline1}
\end{table}

\begin{table}[t]
\centering
\caption{Prompts used in our Pipeline 2.}
{\footnotesize
\renewcommand{\arraystretch}{1.2}
\begin{tabular}{p{0.1\columnwidth}|p{0.8\columnwidth}}
\midrule
\multicolumn{2}{c}{\textbf{Step 2: Augmenting Caption with Negation}} \\
\midrule
\textbf{System} & \texttt{You are a helpful chatbot that generates concise caption.} \\
\hline
\textbf{User} & \texttt{When the answer to the question \{question\} is 'no', reconstruct the caption '\{caption\}'.} \\
\hline
\textbf{LLM} & \texttt{\{updated\_caption\}.} \\
\midrule
\end{tabular}
}
\label{table:prompt_pipeline2}
\end{table}
\section{NegRefCOCOg Benchmark}
\label{sec:negrefcoco}




In this section, we elaborate details on the construction of our proposed NegRefCOCOg benchmark in Sec. 3.3.

\paragraph{Selection Criteria}
To construct NegRefCOCOg, we begin by selecting samples from the RefCOCOg~\cite{refcoco} dataset that meet the following criteria:
\begin{itemize}
    \item The original image patch \( P_o^{+} \), corresponding to the negation-inclusive prompt \( T \), has a height and width of at least 100 pixels.
    \item At least one other image patch belonging to the same category as \( P_o^{+} \) has a height and width of at least 100 pixels and does not overlap with \( P_o^{+} \). 
    We then designate one of these patches as \( P_o^{-} \).
\end{itemize}


\paragraph{Image Patch Maximization}
To ensure alignment with \( T \), we maximize the sizes of \( P_o^{+} \) and \( P_o^{-} \) under the following constraints.
As a result, we obtain the final patches, \( P^{+} \) and \( P^{-} \), where \( P^{+} \) is the expanded version of \( P_o^{+} \) and \( P^{-} \) is the expanded version of \( P^{-} \).

\begin{itemize}
    \item The expanded patch must not overlap with the other patch before its maximization, which can be expressed as:
    \[
    P^{+} \cap P_o^{-} = \varnothing, \quad P^{-} \cap P_o^{+} = \varnothing.
    \]
    \item Horizontal expansion is limited to the original width of the patch in each direction (left and right).
    \item Vertical expansion is limited to the original height of the patch in each direction (top and bottom).
\end{itemize}

\vspace{0.2cm}

We obtain 440 triplets of (\( T \), \( P^{+} \), \( P^{-} \)) through this process, where \( P^{+} \) is well-aligned with \( T \) and \( P^{-} \) serves as challenging hard negative for evaluation.
These 440 samples form the NegRefCOCOg benchmark, which we use to evaluate the ability of models to handle negation comprehensively and accurately.

\begin{figure*}[t]
    \centering
    \includegraphics[width=1.\linewidth]{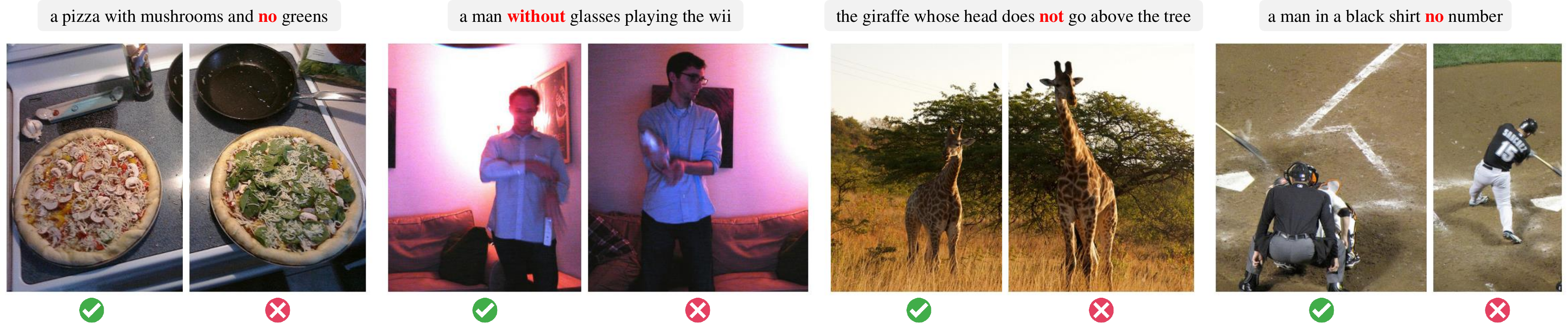}
    \caption{Examples from the proposed NegRefCOCOg benchmark. Each example consists of a textual description containing negation (top) and two image patches: the correct patch aligned with the negated description (green checkmark) and the incorrect patch representing a challenging hard negative (red cross).}
    \label{fig:negrefcoco_examples}
\end{figure*}

\cref{fig:negrefcoco_examples} illustrates examples from NegRefCOCOg. 
These examples demonstrate the diversity of negation scenarios in NegRefCOCOg, including object absence, action negation with various negation terms.

\section{Results on Additional Models}
\label{sec:siglip}

Our method is model-agnostic and applies to SigLIP~\cite{siglip}, showing consistent gains on negation benchmarks while preserving general performance, as shown in \cref{table:siglip}.


\begin{table}[ht]
    \centering
    \caption{Comparison of model performance on negation and general benchmarks across different architectures of SigLIP~\cite{siglip}.}
    \resizebox{\linewidth}{!}{
    \begin{tabular}{lccccc}
        \toprule
        \multirow{2}{*}{\textbf{Model}} & \multirow{2}{*}{\textbf{Arch.}} & \multicolumn{2}{c}{\textbf{Negation Benchmarks}} & \multicolumn{2}{c}{\textbf{General Benchmarks}} \\
        \cmidrule(lr){3-4} \cmidrule(lr){5-6}
                       &                       & \textbf{VALSE} & \textbf{NegRefCOCOg} & \textbf{ImageNet} & \textbf{COCO} \\
        \midrule
        SigLIP & \multirow{2}{*}{base} & 67.98 & 61.59 & \textbf{75.69} & 72.10 \\
        Neg.SigLIP & & \textbf{77.15} & \textbf{68.18} & 75.10 & \textbf{77.30} \\
        \midrule
        SigLIP & \multirow{2}{*}{large} & 73.03 & 62.05 & 79.73 & 75.20 \\
        Neg.SigLIP & & \textbf{79.40} & \textbf{67.73} & \textbf{79.76} & \textbf{80.10} \\
        \midrule
        SigLIP & \multirow{2}{*}{so400m} & 75.47 & 63.18 & \textbf{82.26} & 76.30 \\
        Neg.SigLIP &                     & \textbf{84.27} & \textbf{67.05} & 81.68 & \textbf{81.70} \\
        \bottomrule
    \end{tabular}
    }
    \label{table:siglip}
\end{table}
\section{Expanded Discussion on T2I Generation}
\label{sec:more_t2i}

\subsection{Methods}
In T2I generation,  research has emerged focusing on the removal of unwanted concepts from generated images.~\cite{t2inegation, lmd, sld}.
These methods typically preprocess text prompts by identifying and removing negation-related components, reformulating the prompt to exclude negation.
For example, given the text prompt ``a panda in a forest without flowers,'' LMD~\cite{lmd} uses an LLM is used to identify ``flowers'' as the negated object and remove it, resulting in a final layout-based prompt that retains only ``a panda'' and ``a forest.''

In contrast, our approach directly uses the original prompt ``a panda in a forest without flowers'' without any preprocessing, allowing the T2I model to process negation as part of the input text.
This provides a different perspective on handling negation, where the model learns to interpret negation within natural language rather than relying on preprocessing to modify the prompt.

\subsection{Evaluation}
Existing methods evaluate negation comprehension in T2I models using the LMD~\cite{lmd} Negation benchmark, which consists of 10 prompts structured as ``a realistic photo of a scene without [object]'' and uses object detectors to verify object absence.

\begin{table}[ht]
    \centering
    \caption{Results on LMD Negation benchmark.}
    {
    \footnotesize
    \begin{tabular}{lc}
        \toprule
        \textbf{Method} & \textbf{LMD Negation} \\
        \midrule
        SD &  20\% \\
        SD w/ NegationCLIP text encoder & 94\% \\
        SD + LMD~\cite{lmd} & 100\% \\
        \bottomrule
    \end{tabular}
    }
    \label{table:lmd}
\end{table}

We provide evaluation results of NegationCLIP on the LMD Negation benchmark in \cref{table:lmd}.
Standard SD struggles significantly with negation, achieving only 20\% accuracy, indicating that it often fails to remove the negated object.
In contrast, simply replacing SD's text encoder with NegationCLIP’s text encoder improves accuracy to 94\%, successfully removing the negated object in nearly all cases.
Unsurprisingly, applying LMD~\cite{lmd} method to SD yields perfect performance, as it heuristically removes the negation-related part from the final prompt, ensuring that the object does not appear in the generated image. 

However, while the LMD Negation benchmark effectively evaluates object removal, it does not account for negation beyond object presence or absence, such as actions and attributes.
While it can determine whether an object has been successfully removed, it fails to capture cases where negation modifies an entity’s state (e.g., \textit{a dog not running}) or its properties (e.g., \textit{a not blue sphere}).
To address this limitation, we introduce more diverse prompts that encompass negation in objects, actions, and attributes as shown in \cref{table:prompts_t2i}).

Since object detectors are insufficient for evaluating these more complex negation cases, we drew inspiration from the TIFA~\cite{tifa} metric’s MLLM-based VQA system and introduce the Neg Score as a more comprehensive evaluation measure in Sec 5.2.
\section{Experiment Details}
\label{sec:experiment_details}


\subsection{CelebA Classification}

In \cref{table:celeba_prompts}, we provide attribute-specific prompts for all 40 attributes of CelebA~\cite{celeba} in our experiment in Sec. 2.1.
NegationCLIP improves the accuracy from 60.8\% to 64.0\%, whereas CoN-CLIP~\cite{conclip} sees a decline to 60.6\%.
We note that achieving significantly higher accuracy on CelebA is inherently challenging, as the dataset includes subjective attributes (e.g., \textit{attractive}).

\begin{table*}[t]
    \centering
    \caption{CelebA attribute-specific prompts and balanced accuracy}
    {\footnotesize
    \renewcommand{\arraystretch}{1.2} 
    \begin{tabular}{l|l|l}
        \toprule
        \textbf{Attribute} & \textbf{Positive Prompt} & \textbf{Negative Prompt} \\
        \midrule
        5 o Clock Shadow & {a photo of a person with a 5 o'clock shadow} & {a photo of a person with no 5 o'clock shadow} \\
        Arched Eyebrows & {a photo of a person with arched eyebrows} & {a photo of a person with not arched eyebrows} \\
        Attractive & {a photo of an attractive person} & {a photo of a not attractive person} \\
        Bags Under Eyes & {a photo of a person with bags under eyes} & {a photo of a person with no bags under eyes} \\
        Bald & {a photo of a bald person} & {a photo of a not bald person} \\
        Bangs & {a photo of a person with bangs} & {a photo of a person with no bangs} \\
        Big Lips & {a photo of a person with big lips} & {a photo of a person with not big lips} \\
        Big Nose & {a photo of a person with a big nose} & {a photo of a person with a not big nose} \\
        Black Hair & {a photo of a person with black hair} & {a photo of a person with not black hair} \\
        Blond Hair & {a photo of a person with blond hair} & {a photo of a person with not blond hair} \\
        Blurry & {a blurry photo of a person} & {a not blurry photo of a person} \\
        Brown Hair & {a photo of a person with brown hair} & {a photo of a person with not brown hair} \\
        Bushy Eyebrows & {a photo of a person with bushy eyebrows} & {a photo of a person with not bushy eyebrows} \\
        Chubby & {a photo of a chubby person} & {a photo of a not chubby person} \\
        Double Chin & {a photo of a person with a double chin} & {a photo of a person with no double chin} \\
        Eyeglasses & {a photo of a person wearing glasses} & {a photo of a person not wearing glasses} \\
        Goatee & {a photo of a person with goatee} & {a photo of a person with no goatee} \\
        Gray Hair & {a photo of a person with gray hair} & {a photo of a person with not gray hair} \\
        Heavy Makeup & {a photo of a person with heavy makeup} & {a photo of a person with no heavy makeup} \\
        High Cheekbones & {a photo of a person with high cheekbones} & {a photo of a person with not high cheekbones} \\
        Male& {a photo of a male} & {a photo of a not male} \\
        Mouth Slightly Open & {a photo of a person with mouth slightly open} & {a photo of a person with mouth not slightly open} \\
        Mustache & {a photo of a person with mustache} & {a photo of a person with no mustache} \\
        Narrow Eyes & {a photo of a person with narrow eyes} & {a photo of a person with not narrow eyes} \\
        No Beard & {a photo of a person with no beard} & {a photo of a person with beard} \\
        Oval Face & {a photo of a person with oval face} & {a photo of a person with not oval face} \\
        Pale Skin & {a photo of a person with pale skin} & {a photo of a person with not pale skin} \\
        Pointy Nose & {a photo of a person with a pointy nose} & {a photo of a person with not a pointy nose} \\
        Receding Hairline & {a photo of a person with a receding hairline} & {a photo of a person with no receding hairline} \\
        Rosy Cheeks & {a photo of a person with rosy cheeks} & {a photo of a person with not rosy cheeks} \\
        Sideburns & {a photo of a person with sideburns} & {a photo of a person with no sideburns} \\
        Smiling& {a photo of a person smiling} & {a photo of a person not smiling} \\
        Straight Hair & {a photo of a person with straight hair} & {a photo of a person with not straight hair} \\
        Wavy Hair & {a photo of a person with wavy hair} & {a photo of a person with not wavy hair} \\
        Wearing Earrings& {a photo of a person wearing earrings} & {a photo of a person not wearing earrings} \\
        Wearing Hat & {a photo of a person wearing a hat} & {a photo of a person not wearing a hat} \\
        Wearing Lipstick & {a photo of a person wearing lipstick} & {a photo of a person not wearing lipstick} \\
        Wearing Necklace & {a photo of a person wearing a necklace} & {a photo of a person not wearing a necklace} \\
        Wearing Necktie & {a photo of a person wearing a necktie} & {a photo of a person not wearing a necktie} \\
        Young & {a photo of a young person} & {a photo of a not young person} \\
        \bottomrule
    \end{tabular}
    }
    \label{table:celeba_prompts}
\end{table*}

\subsection{Fine-Tuning Configuration}

We detail the fine-tuning configurations used to train our CLIP~\cite{clip} models with the data generated from the proposed data generation pipeline.

The generated dataset was split into 80\% for training and 20\% for validation.
We used batch sizes of 512 for ViT-B/32, 256 for ViT-B/16, 128 for both ViT-L/14 and ViT-L@336px, and 64 for ViT-BigG/14 which we fine-tuned specifically for the text-to-image (T2I) generation experiment using SDXL-1.0~\cite{sdxl} in Sec. 5.1.
All models were fine-tuned using a single NVIDIA L40 GPU.

\subsection{T2I Generation}

\begin{table*}
\centering
\caption{Text-to-image generation prompts and corresponding questions.}
\resizebox{2.\columnwidth}{!}
{\scriptsize
\renewcommand{\arraystretch}{1.2} 
\begin{tabular}{p{0.19\textwidth}|p{0.12\textwidth}|p{0.23\textwidth}||p{0.19\textwidth}|p{0.12\textwidth}|p{0.23\textwidth}}
\toprule
\textbf{Prompt} & \textbf{Question 1} & \textbf{Question 2} & \textbf{Prompt} & \textbf{Question 1} & \textbf{Question 2} \\
\midrule
a man not wearing a hat & Is this a man? & Is the man wearing a hat? & a woman not wearing a mask & Is this a woman? & Is the woman wearing a mask? \\
a dog not running & Is this a dog? & Is the dog running? & a bird not flying & Is this a bird? & Is the bird flying? \\
a not white cat & Is this a cat? & Is the cat white? & a not blue sphere & Is this a sphere? & Is the sphere blue? \\
a room with no window & Is this a room? & Is there a window in the room? & a dog without a collar & Is this a dog? & Does the dog have a collar? \\
a cat with no whiskers & Is this a cat? & Does the cat have whiskers? & a child not holding a toy & Is this a child? & Is the child holding a toy? \\
a park with no benches & Is this a park? & Are there benches in the park? & a bird not perched on a branch & Is this a bird? & Is the bird perched on a branch? \\
a woman without glasses & Is this a woman? & Is the woman wearing glasses? & a table with no chairs around & Is this a table? & Are there chairs around the table? \\
a car not parked in the driveway & Is this a car? & Is the car parked in the driveway? & a beach without any umbrellas & Is this a beach? & Are there umbrellas on the beach? \\
a man with no hat & Is this a man? & Is the man wearing a hat? &
a road not crowded with cars & Is this a road? & Is the road crowded with cars? \\
a garden with no flowers & Is this a garden? & Are there flowers in the garden? &
a person not holding an umbrella & Is this a person? & Is the person holding an umbrella? \\
a lake with no boats & Is this a lake? & Are there boats on the lake? &
a house without a roof & Is this a house? & Does the house have a roof? \\
a tree not in bloom & Is this a tree? & Is the tree in bloom? &
a mountain with no snow & Is this a mountain? & Is there snow on the mountain? \\
a room without furniture & Is this a room? & Is there furniture in the room? &
a dog not barking & Is this a dog? & Is the dog barking? \\
a street with no people & Is this a street? & Are there people on the street? &
a kitchen without any food & Is this a kitchen? & Is there food in the kitchen? \\
a cup not filled with coffee & Is this a cup? & Is the cup filled with coffee? &
a forest with no animals & Is this a forest? & Are there animals in the forest? \\
a phone not on the table & Is this a phone? & Is the phone on the table? &
a desk without a computer & Is this a desk? & Is there a computer on the desk? \\
a man not wearing shoes & Is this a man? & Is the man wearing shoes? &
a restaurant with no tables & Is this a restaurant? & Are there tables in the restaurant? \\
a city skyline with no skyscrapers & Is this a city skyline? & Are there skyscrapers in the city skyline? &
a field without crops & Is this a field? & Are there crops in the field? \\
a woman not smiling & Is this a woman? & Is the woman smiling? &
a living room with no couch & Is this a living room? & Is there a couch in the living room? \\
a car without wheels & Is this a car? & Does the car have wheels? &
a stadium with no spectators & Is this a stadium? & Are there spectators in the stadium? \\
a road with no signs & Is this a road? & Are there signs on the road? &
a child not wearing shoes & Is this a child? & Is the child wearing shoes? \\
a bridge without railings & Is this a bridge? & Does the bridge have railings? &
a river with no fish & Is this a river? & Are there fish in the river? \\
a sky without clouds & Is this a sky? & Are there clouds in the sky? &
a cup with no handle & Is this a cup? & Does the cup have a handle? \\
a playground with no swings & Is this a playground? & Are there swings in the playground? &
a man not wearing a tie & Is this a man? & Is the man wearing a tie? \\
a building without windows & Is this a building? & Does the building have windows? &
a book with no cover & Is this a book? & Does the book have a cover? \\
a shop with no customers & Is this a shop? & Are there customers in the shop? &
a garden without any trees & Is this a garden? & Are there trees in the garden? \\
a bike not leaning against a wall & Is this a bike? & Is the bike leaning against a wall? &
a stage with no performers & Is this a stage? & Are there performers on the stage? \\
a train station with no trains & Is this a train station? & Are there trains at the train station? &
a museum without exhibits & Is this a museum? & Are there exhibits in the museum? \\
a shelf with no books & Is this a shelf? & Are there books on the shelf? &
a restaurant not serving food & Is this a restaurant? & Is the restaurant serving food? \\
a person with no backpack & Is this a person? & Does the person have a backpack? &
a market without any vendors & Is this a market? & Are there vendors in the market? \\
a room not filled with light & Is this a room? & Is the room filled with light? &
a path with no signs & Is this a path? & Are there signs on the path? \\
a school without students & Is this a school? & Are there students in the school? &
a car with no headlights & Is this a car? & Does the car have headlights? \\
a cat without a tail & Is this a cat? & Does the cat have a tail? &
a person not holding a bag & Is this a person? & Is the person holding a bag? \\
a forest with no leaves & Is this a forest? & Are there leaves in the forest? &
a house with no doors & Is this a house? & Does the house have doors? \\
a chair not facing the table & Is this a chair? & Is the chair facing the table? &
a bird not singing & Is this a bird? & Is the bird singing? \\
a beach without sand & Is this a beach? & Is there sand on the beach? &
a dog not playing fetch & Is this a dog? & Is the dog playing fetch? \\
a wall with no decorations & Is this a wall? & Are there decorations on the wall? &
a sidewalk with no pedestrians & Is this a sidewalk? & Are there pedestrians on the sidewalk? \\
a man not reading a book & Is this a man? & Is the man reading a book? &
a classroom with no desks & Is this a classroom? & Are there desks in the classroom? \\
a street with no lights & Is this a street? & Are there lights on the street? &
a yard without grass & Is this a yard? & Is there grass in the yard? \\
a riverbank with no trees & Is this a riverbank? & Are there trees on the riverbank? &
a cat not purring & Is this a cat? & Is the cat purring? \\
a boat with no sails & Is this a boat? & Does the boat have sails? &
a woman not holding a purse & Is this a woman? & Is the woman holding a purse? \\
a stadium with no players & Is this a stadium? & Are there players in the stadium? &
a sky with no stars & Is this a sky? & Are there stars in the sky? \\
a store without shelves & Is this a store? & Are there shelves in the store? &
a man not holding a briefcase & Is this a man? & Is the man holding a briefcase? \\
a city without buildings & Is this a city? & Are there buildings in the city? &
a painting without colors & Is this a painting? & Does the painting have colors? \\
a road without any turns & Is this a road? & Are there turns on the road? &
a lawn with no flowers & Is this a lawn? & Are there flowers on the lawn? \\
a dog not fetching a ball & Is this a dog? & Is the dog fetching a ball? &
a bridge without any lights & Is this a bridge? & Are there lights on the bridge? \\
a car with no passengers & Is this a car? & Are there passengers in the car? &
a garden with no vegetables & Is this a garden? & Are there vegetables in the garden? \\
a child not drinking milk & Is this a child? & Is the child drinking milk? &
a person without a shadow & Is this a person? & Does the person have a shadow? \\
a tree with no leaves & Is this a tree? & Does the tree have leaves? &
a bus stop without a bench & Is this a bus stop? & Is there a bench at the bus stop? \\
a train without passengers & Is this a train? & Are there passengers on the train? &
a cafe with no tables & Is this a cafe? & Are there tables in the cafe? \\
a photo with no people & Is this a photo? & Are there people in the photo? &
a street without any trees & Is this a street? & Are there trees on the street? \\
a river not flowing & Is this a river? & Is the river flowing? &
a mountain with no trails & Is this a mountain? & Are there trails on the mountain? \\
a path without any footprints & Is this a path? & Are there footprints on the path? &
a field with no animals & Is this a field? & Are there animals in the field? \\
a building with no entrance & Is this a building? & Is there an entrance to the building? \\
\bottomrule
\end{tabular}
}
\label{table:prompts_t2i}
\end{table*}

We provide details on T2I experiments shown in Sec. 5.1.

For generating negation-inclusive prompts, we utilized ChatGPT~\cite{chatgpt} to construct 107 prompts.
To evaluate whether the generated images accurately reflected the given prompts, we employed ChatGPT to create two corresponding questions for each prompt.
The first question was designed to assess whether the subject of the prompt was correctly generated, with the expected answer being ``yes.'' The second question aimed to verify whether the negation-related aspect was properly removed, with the expected answer being ``no.''
For example, for the prompt ``a man not wearing a hat,'' the two generated questions were ``Is this a man?'' and ``Is the man wearing a hat?''.
The complete list of prompts and their corresponding questions is provided in ~\cref{table:prompts_t2i}.

In SD-1.4~\cite{sd1.4}, we replaced its CLIP ViT-L/14 text encoder with CoN-CLIP text encoder and our fine-tuned negation-aware text encoder of the same architecture. 
For SDXL, which employs two text encoders (CLIP ViT-L/14 and CLIP ViT-BigG/14), we substituted both encoders with our fine-tuned versions.

\subsection{Referring Image Segmentation}

For our experiments in referring image segmentation in Sec. 5.2, we utilized the publicly available weight of CLIPSeg~\cite{clipseg} trained on the PhraseCut~\cite{phrasecut} dataset as our baseline. 
Without any additional training, we replaced the text encoder in the CLIPSeg architecture with CoN-CLIP ViT-B/16 text encoder and our fine-tuned CLIP ViT-B/16 text encoder.

We followed CLIPSeg to determine the threshold values for binary segmentation, setting it to 0.3 for experiments on the PhraseCut dataset and 0.1 for experiments on RefCOCOg (Neg), which is based on the COCO~\cite{coco} dataset.

\section{Additional Qualitative Results}
\label{sec:qualitative_results}

\begin{figure*}[t]
    \centering
    \includegraphics[width=1.\linewidth]{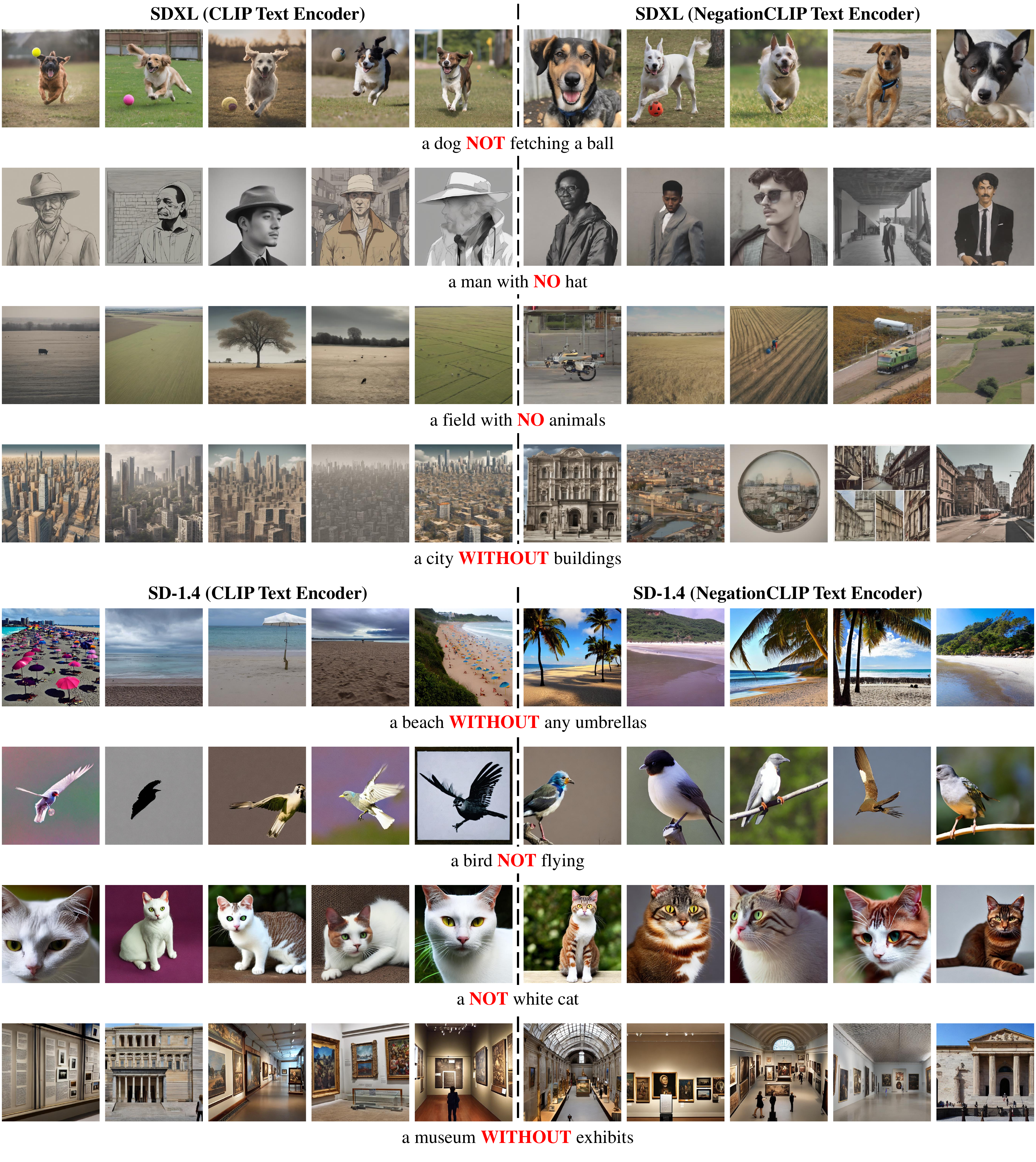}
    \caption{Additional examples for text-to-image generation tasks using Stable Diffusion XL (SDXL)~\cite{sdxl} and Stable Diffusion 1.4 (SD-1.4)~\cite{sd1.4}. Comparisons are shown between models using the original CLIP text encoder and the fine-tuned NegationCLIP text encoder.}
    \label{fig:t2i_appen}
\end{figure*}

\cref{fig:t2i_appen} demonstrates additional qualitative examples from T2I task.
For each prompt, we include all images generated using 5 different random seeds.

For both SDXL~\cite{sdxl} and SD-1.4~\cite{sd1.4}, substituting the original CLIP text encoder with NegationCLIP text encoder improves the ability to accurately reflect negation in generated images.
Notably, SD-1.4, which uses only one text encoder, maintains high-quality image generation despite the substitution, highlighting that the fine-tuned NegationCLIP text encoder preserves overall image quality while enhancing negation comprehension.

Certain challenging prompts, such as ``a city without buildings'' or ``a museum without exhibits,'' expose limitations in the model's ability to remove concepts with strong biases (e.g., buildings in city contexts, exhibits in museum contexts). 
These limitations may stem from the limited representation of such cases in the training data for generative models.
Identifying the underlying causes of these limitations and addressing them constitutes an important direction for future work.




\end{document}